\documentclass[letterpaper, 10 pt, conference]{ieeeconf}  

\IEEEoverridecommandlockouts                              

\usepackage{dblfloatfix} 
\usepackage[utf8]{inputenc} 
\usepackage[T1]{fontenc}    
\usepackage{hyperref}       
\usepackage{url}            
\usepackage{booktabs}       
\usepackage{amsfonts}       
\usepackage{nicefrac}       
\usepackage{microtype}      
\usepackage{amsmath}
\usepackage{subcaption}
\usepackage{wrapfig}
\usepackage{gensymb}

\usepackage[font=small,labelfont=bf]{caption}

\usepackage{graphics} 
\usepackage{epsfig} 
\usepackage{amssymb}  
\usepackage{wrapfig}
\usepackage{authblk}
\title{Dexterous Manipulation with Deep Reinforcement Learning:\\Efficient, General, and Low-Cost}

\author{Henry Zhu$^*$${^1}$\thanks{$^*$The first two authors contributed equally to this work.} \hspace{3mm} Abhishek Gupta$^*$${^1}$ \hspace{3mm} Aravind Rajeswaran${^2}$ \hspace{3mm} Sergey Levine${^1}$ \hspace{3mm} Vikash Kumar${^3}$ \\
$^1$ UC Berkeley \hspace{0.2in} $^2$ University of Washington \hspace{0.2in} $^3$ Google Brain\\
}

\newcommand{\cM}{\mathcal{M}}
\newcommand{\cD}{\mathcal{D}}

\newcommand{\bR}{\mathbb{R}}
\newcommand{\bE}{\mathbb{E}}

\newcommand{\pith}{\pi_\theta}

\begin{document}

\maketitle

\begin{abstract}
Dexterous multi-fingered robotic hands can perform a wide range of manipulation skills, making them an appealing component for general-purpose robotic manipulators. However, such hands pose a major challenge for autonomous control, due to the high dimensionality of their configuration space and complex intermittent contact interactions. In this work, we propose deep reinforcement learning (deep RL) as a scalable solution for learning complex, contact rich behaviors with multi-fingered hands. Deep RL provides an end-to-end approach to directly map sensor readings to actions, without the need for task specific models or policy classes. We show that contact-rich manipulation behavior with multi-fingered hands can be learned by directly training with model-free deep RL algorithms in the real world, with minimal additional assumption and without the aid of simulation. We learn a variety of complex behaviors on two different low-cost hardware platforms. We show that each task can be learned entirely from scratch, and further study how the learning process can be further accelerated by using a small number of human demonstrations to bootstrap learning. Our experiments demonstrate that complex multi-fingered manipulation skills can be learned in the real world in about 4-7 hours for most tasks, and that demonstrations can decrease this to 2-3 hours, indicating that direct deep RL training in the real world is a viable and practical alternative to simulation and model-based control. \url{https://sites.google.com/view/deeprl-handmanipulation}
\end{abstract}

\section{Introduction}

A long standing goal in robotics is to create general purpose robotic systems that operate in a wide variety of human-centric environments such as homes and hospitals. For robotic agents to be competent in such unstructured environments, versatile manipulators like multi-fingered hands are needed in order to cope with the diversity of tasks presented by human-centric settings. However, the versatility of multi-fingered robotic hands comes at the price of high dimensional configuration spaces and complex finger-object contact interactions, which makes modeling and controller synthesis particularly challenging.  

\begin{figure}[!t]
    \centering
    \includegraphics[width=0.48\columnwidth]{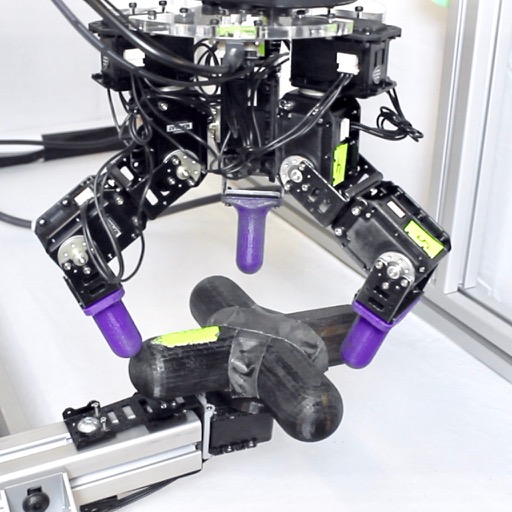}
    \includegraphics[width=0.48\columnwidth]{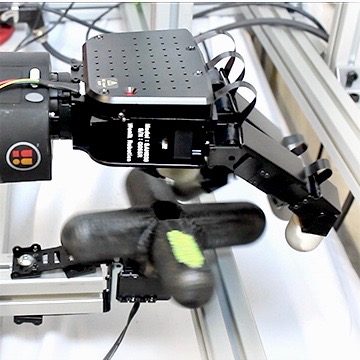}
    \\
     \vspace{.1cm}
    \includegraphics[width=0.32\columnwidth]{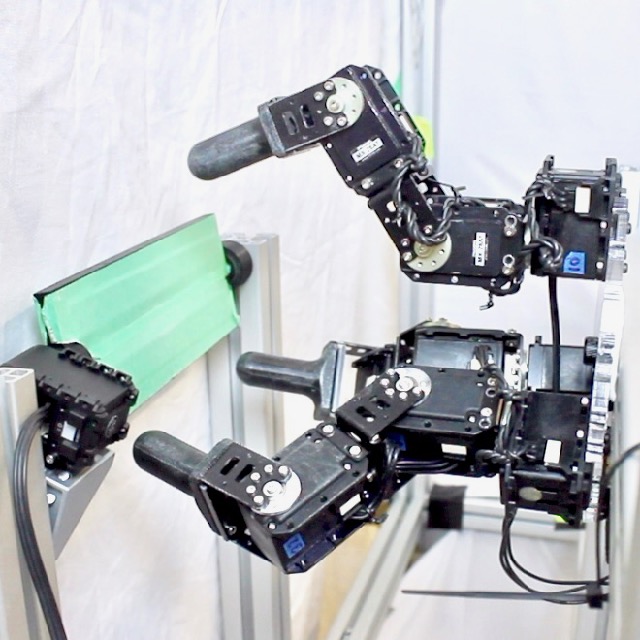}
    \includegraphics[width=0.32\columnwidth]{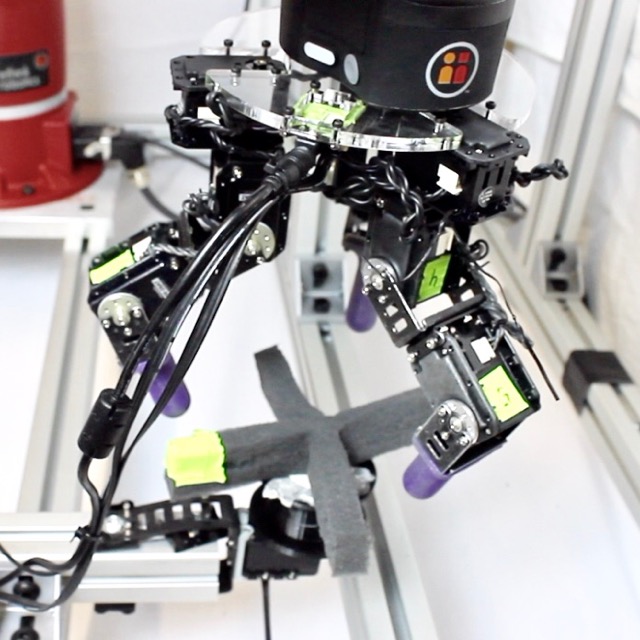}
    \includegraphics[width=0.32\columnwidth, height=0.32\columnwidth]{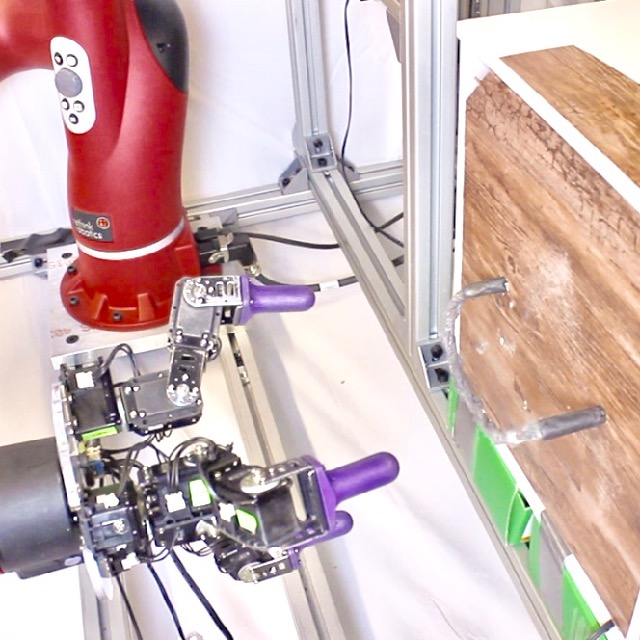}
    
    \caption{We demonstrate that DRL can learn a wide range of dexterous manipulation skills with multi-fingered hands, such as opening door with flexible handle, rotating a cross-shaped valve, and rotating the same valve but with a deformable foam handle, which presents an additional physical challenge, and box flipping.}
    \vspace{-0.5cm}
    \label{fig:behaviors}
\end{figure}

Reinforcement learning (RL) and optimal control techniques provide generic paradigms for optimizing complex controllers that are hard to specify by hand. In particular, model-free RL provides a way to optimize controllers end-to-end without any explicit modeling or system identification. This reduces human engineering effort and produces controllers that are directly adapted to the physical environment, which makes for a scalable and general approach to robotic manipulation: a large repertoire of skills can be acquired by directly learning in different situations in the real world with only the task reward specified by the user. Such systems do not require an explicit model of the robot, calibration of the sensors, or other manual characterization that is typically needed for model-based methods.

In this work, we study how model-free RL can be scaled up to learn a variety of manipulation behaviors with multi-fingered hands directly in the real world, using general-purpose neural network policy representations and without manual controller or policy class design. We conduct our experiments with low cost multi-fingered manipulators, and show results on tasks such as rotating a valve, flipping a block vertically, and door opening. Somewhat surprisingly, we find that successful neural network controllers for each of these tasks can be trained directly in the real world in about 4-7 hours for most tasks. 

We also show that we can further accelerate the learning process by incorporating a small number of human demonstrations, building on the recently proposed DAPG~\cite{DAPG} algorithm. Using 20 demonstrations obtained through kinesthetic teaching, the learning time can be brought down to around $2-3$ hours which corresponds to a 2x speedup. Together, these results establish that model-free deep RL and demonstration-driven acceleration provide a viable approach for real-world learning of diverse manipulation skills.

The main contribution of this work is to demonstrate that model-free deep reinforcement learning can learn contact rich manipulation behaviors directly for low-cost multi-fingered hands directly in the real world. We demonstrate this on two different low cost robotic hands, manipulating both rigid and deformable objects, and performing three distinct tasks. We further demonstrate that a small number of demonstrations can accelerate learning, and analyze a number of design choices and performance parameters.

\section{Related Work}
\label{sec:related}

In recent years, multi-fingered robotic hands have become more and more widespread as robotic manipulation has progressed towards more challenging tasks. A significant portion of this research has focused on the physical design of anthropomorphic hands, and designing simple controllers for these hands. The majority of these manipulators~\cite{adroitvikash, dlrhand} are custom, expensive, heavily instrumented, and fragile. In contrast, the results presented in this work are on low-cost (and relatively more robust) commodity hardware, with limited sensing and actuation capabilities. As our method trains directly in the real world, the resulting solution encapsulates the sensory and control inaccuracies that are associated with such hardware thereby making them more effective and deployable.

Much of the work on robotic hands has focused on grasping~\cite{bicchi2000robotic}, as compared to other dexterous skills~\cite{okamura2000overview}. Such work is often focused on achieving stable grasps by explicitly reasoning about the stability of the grasp using geometric and analytic metrics~\cite{bullock2013hand,zhu2003synthesis,miller2004graspit}. While considerable progress has been made on the theoretical side~\cite{murray2017mathematical}, sensing and estimation challenges (visual occlusions and limited tactile sensing) has limited their applicability in the real world considerably. In this work, we consider tasks which require significant finger coordination and dexterity, making the manual design of controllers very challenging. Our proposed solution using model-free deep RL alleviates the need for manual controller design or modeling. 


In this work, we study the use of model-free deep reinforcement learning algorithms to learn manipulation policies for dexterous multi-finger hands. In simulation,~\cite{mordatch2012contact, bai2014dexterous, mpcvikash} have shown the ability to synthesize complex behaviors when provided with a ground truth model. This model is rarely available in the real world. Techniques such as ~\cite{kumaroptimal}, ~\cite{softhand} get around this problem by learning locally-linear models and deploying model-based RL method to learn instance specific linear controllers demonstrating turning rods and arranging with beads. \cite{vanHoof2015learning} show the ability to learn in-hand repositioning in the real world, with a reinforcement learning method (NPREPS) and non-parametric control policies. Similar methods have also been used for whole-arm manipulation~\cite{petersrobot}. While these methods train in the real world, the resulting policies involve relatively simple individual motions. In contrast to these model-free experiments, our results show that deep RL can learn complex contact-rich behaviors with finger gaits, including with combined hand and arm control. In contrast to the model-based methods, we show that this can be done directly in the real world without any model.

We also show how the learning process can be accelerated using a small number of human demonstrations. Accelerating reinforcement learning with demonstrations has been the subject of a number of works in the reinforcement learning community ~\cite{ijspeert2002learning,petersrobot,DDPGfD, ashvinexploration, DAPG, DQfD, yukeimitation} but these methods have not been applied to real world manipulation with multi-fingered hands. We show that this is indeed possible and very effective using algorithms which combine behavior cloning and policy gradient. We build on the algorithm proposed by us in prior work~\cite{DAPG}, and show that this approach can indeed scale to real world dexterous manipulation, with significant acceleration benefits.

Another related line of research seeks to transfer policies trained in simulation into the real world. Aside from rigorous system identification, a recent class of methods has focused on randomizing the simulation, both for physical~\cite{epopt} and visual~\cite{cad2rl} transfer. This approach has been employed for manipulation~\cite{johns}, visual servoing~\cite{fereshtehservo}, locomotion~\cite{sim2realcheetah}, and navigation~\cite{cad2rl}. Recent concurrent work has also applied this approach to multi-fingered hands for an object rotation task~\cite{dexterousopenai} with about 50 hours of computation, equivalent to 100 years of experience. In contrast, our approach trains directly in the real world in a few hours without the need for manual modeling. We discuss the relative merits of real-world and simulated training in Section~\ref{sec:sim2real}.

\section{Hardware Setup}
In order to demonstrate the generalizable nature of the model-free deep RL algorithms, we consider two different hardware platforms: a custom built 3 fingered hand, referred to as the Dynamixel claw (Dclaw), and a 4 fingered Allegro hand. Both hands are relatively cheap, especially the Dclaw, which costs under $\$2,500$ to build. For several experiments, we also mounted the hands on a Sawyer robot arm to allow for a larger workspace.

\paragraph{Dynamixel Claw}
The Dynamixel claw (Dclaw) is custom built using Dynamixel servo motors. It is a powerful, low latency, position controlled 9 DoF manipulator which costs under $\$2,500$ to construct. Dclaw is robust and is able to run up to 24 hours without intervention or hardware damage. 

\paragraph{Allegro Hand}
The Allegro hand is a 4 fingered anthropomorphic hand, with 16 degrees of freedom, and can handle payloads of up to 5 kg. This hand uses DC motors for actuation and can be either torque or position controlled using a low level PID. The Allegro hand costs on the order of $\$15,000$.

\begin{figure}[!h]
  \centering
  \includegraphics[width= 0.4\columnwidth]{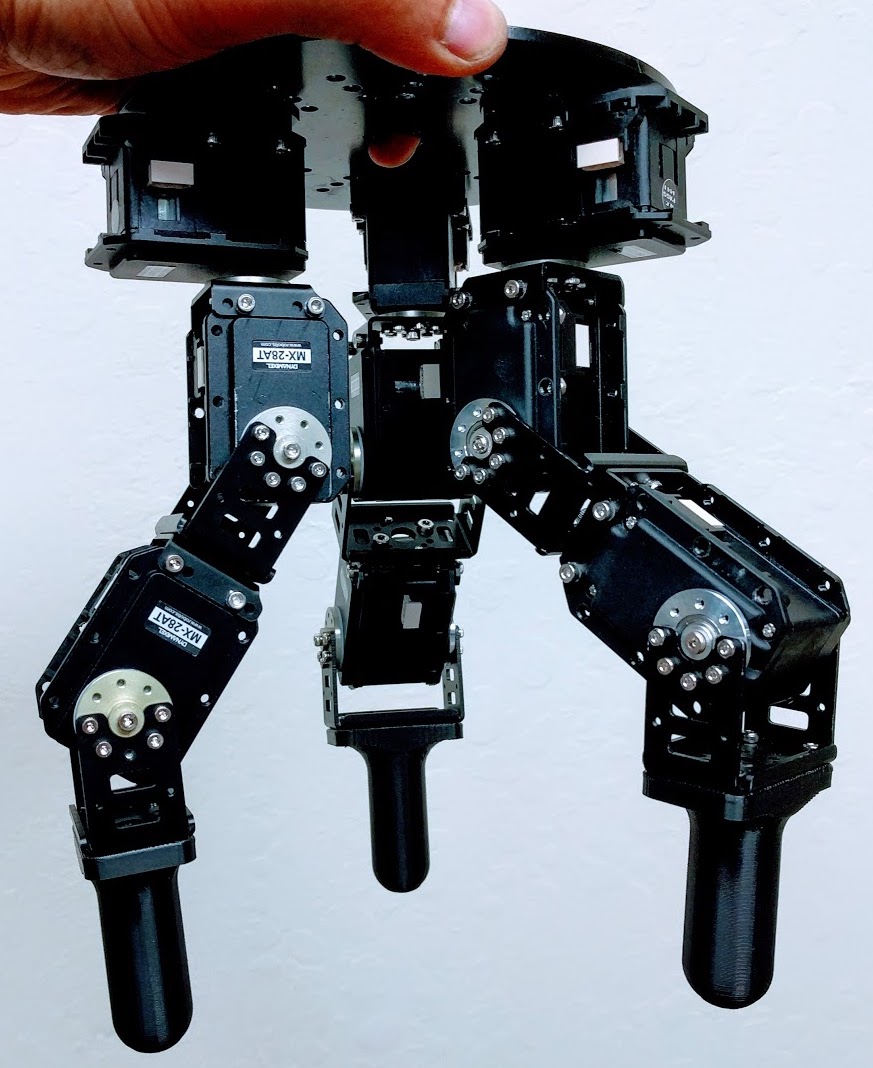}
  \includegraphics[width= 0.39\columnwidth]{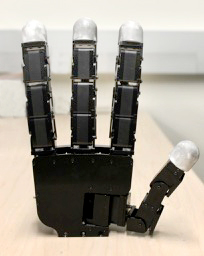}
  \caption{Left: 3 finger Dynamixel claw. Right: 4 finger anthropomorphic Allegro hand}
  \label{fig:robots}
\end{figure} 

\section{Tasks}
\label{sec:tasks}

While the approach we describe for learning dexterous manipulation is general and broadly applicable, we consider three distinct tasks in our experimental evaluation - valve rotation, box flipping, and door opening. These tasks involve challenging contact patterns and coordination, and are inspired by everyday hand manipulations. We approach these problems using reinforcement learning, modeling them as Markov decision processes (MDPs), which provide a generic mathematical abstraction to model sequential decision making problems. The goal in reinforcement learning is to learn a control policy which maximizes a user-provided reward function. This reward function is defined independently for each of our tasks as described below.

\paragraph{Valve Rotation}
This task involves turning a valve or faucet to a target position. The fingers must cooperatively push and move out of the way, posing an exploration challenge. Furthermore the contact forces with the valve complicate the dynamics. For our task, the valve must be rotated from  $0\degree$ to $180\degree$.

\begin{figure}[!h]
  \centering
  \includegraphics[origin=c,width=0.23\columnwidth]{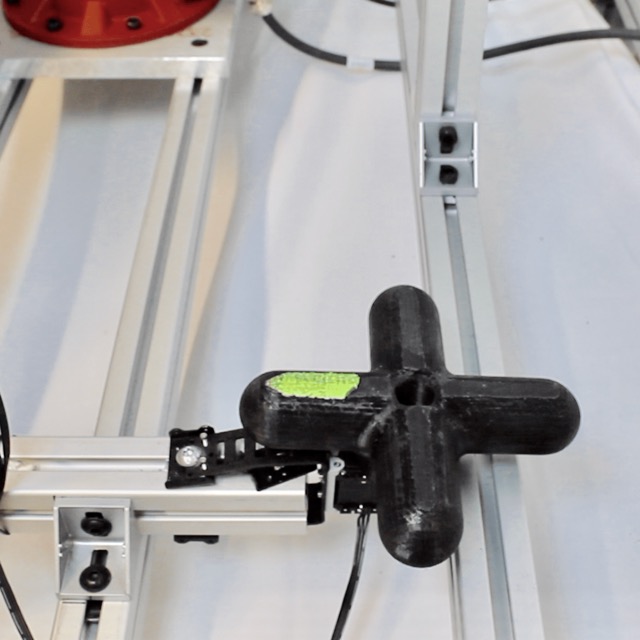}
  \includegraphics[origin=c,width=0.23\columnwidth]{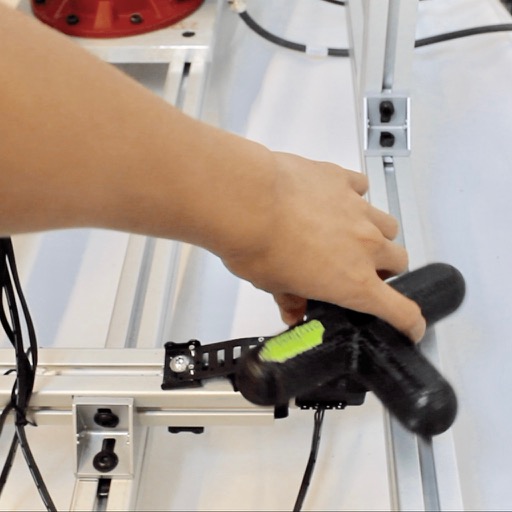}
  \includegraphics[origin=c,width=0.23\columnwidth]{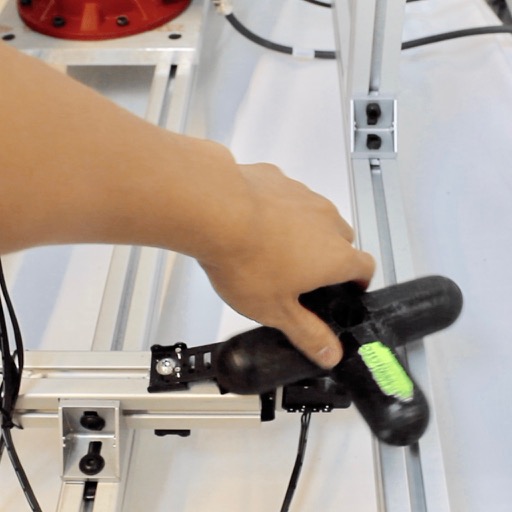}
  \includegraphics[origin=c,width=0.23\columnwidth]{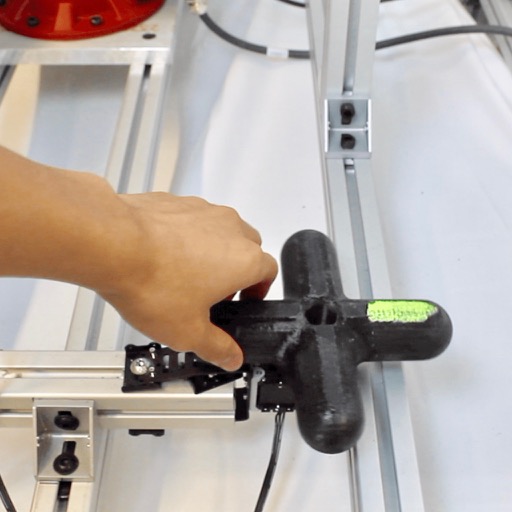}
  \caption{Illustration of valve rotation}
  \label{fig:valve_rotation}
\end{figure} 

The state space consists of all the joint angles of the hand, the current angle of rotation of the valve [$\theta_{\text{valve}}$], the distance to the goal angle [$d\theta$], and the last action taken. The action space is joint angles of the hand and the reward function is 
\begin{gather*}
    r = -|d\theta| + 10*\mathbb{1}_{\{|d\theta| < 0.1\}} + 50*\mathbb{1}_{\{|d\theta| < 0.05\}} \\
    d\theta := \theta_{\text{valve}} - \theta_{\text{goal}}
\end{gather*}

We define a trajectory as a success if $|d\theta| < 20\degree$ for at least $20\%$ of the trajectory.

\paragraph{Vertical box flipping}
This task involves rotating a rectangular box, which freely spins about its long axis, from $0\degree$ to $180\degree$. This task also involves learning alternating coordinated motions of the fingers such that while the top finger is pushing, the bottom two move out of the way, and vice versa.

\begin{figure}[!h]
  \centering
  \includegraphics[origin=c,width=0.23\columnwidth]{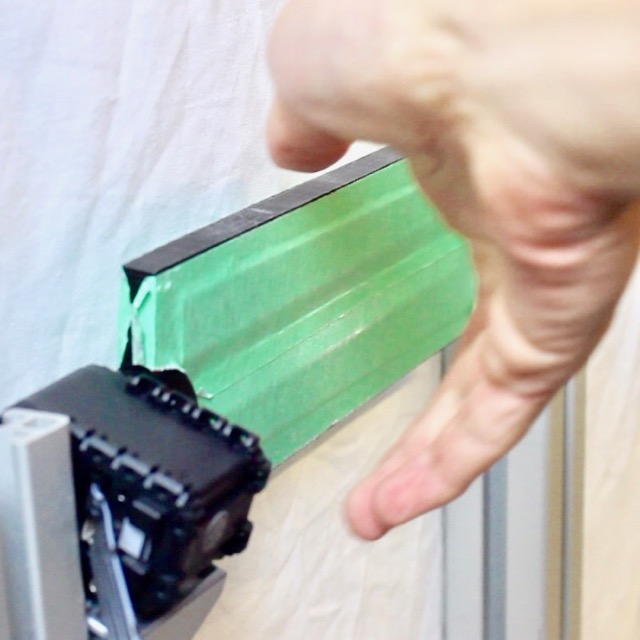}
  \includegraphics[origin=c,width=0.23\columnwidth]{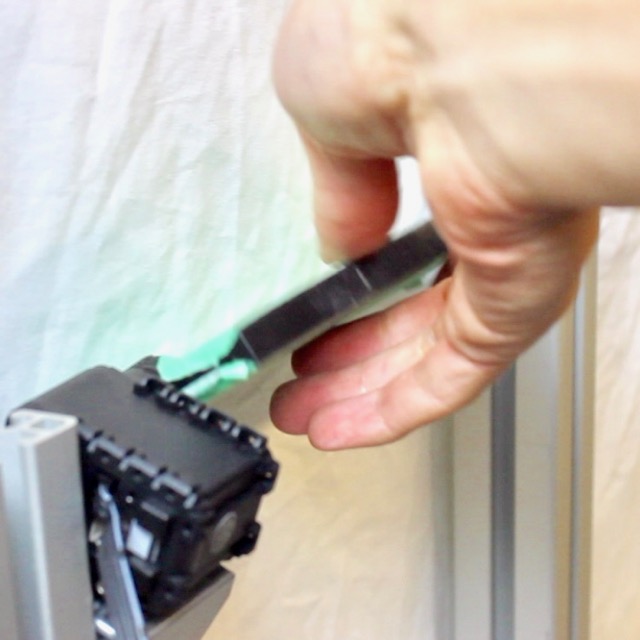}
  \includegraphics[origin=c,width=0.23\columnwidth]{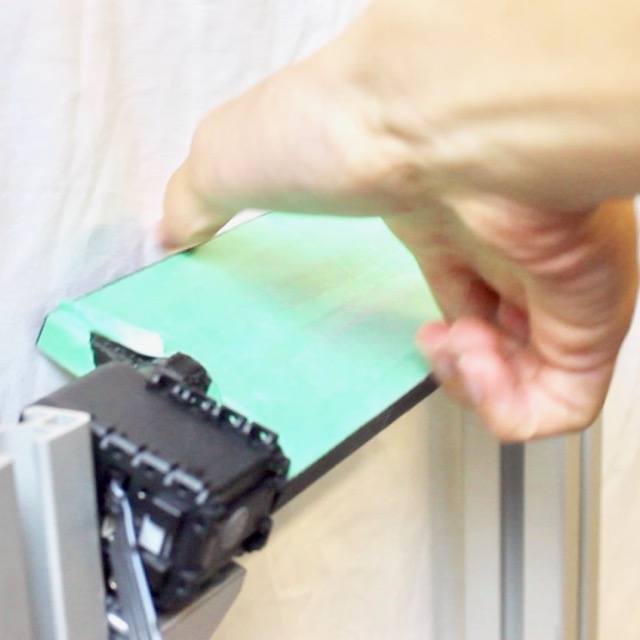}
  \includegraphics[origin=c,width=0.23\columnwidth]{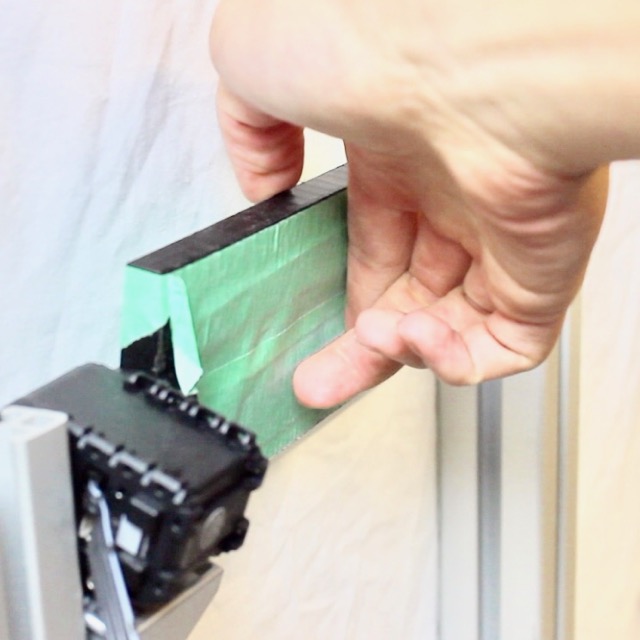}
  \caption{Illustration of box flipping}
  \label{fig:flipping_hand}
\end{figure}

The state space consists of all the joint angles of the hand, the current angle of rotation of the box [$\theta_{\text{box}}$], the distance in angle to the goal, and the last action taken. The action space consists of the joint angles of the hand and the reward function is 
\begin{gather*}
    r = -|d\theta| + 10*\mathbb{1}_{\{|d\theta| < 0.1\}} + 50*\mathbb{1}_{\{|d\theta| < 0.05\}} \\
    d\theta := \theta_{\text{box}} - \theta_{\text{goal}}
\end{gather*}

We define a trajectory as a success if $|d\theta| < 20\degree$ for at least $20\%$ of the trajectory.

\paragraph{Door opening}
This task involves both the arm and the hand working in tandem to open a door. The robot must learn to approach the door, grip the handle, and then pull backwards. This task has more degrees of freedom given the additional arm, and involves the sequence of actions: going to the door, gripping the door, and then pulling away.

\begin{figure}[!h]
  \centering
  \includegraphics[origin=c,width=0.23\columnwidth]{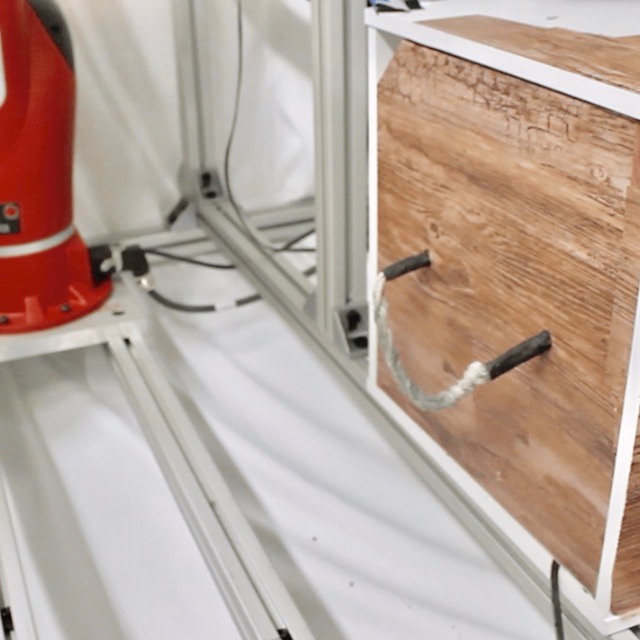}
  \includegraphics[origin=c,width=0.23\columnwidth]{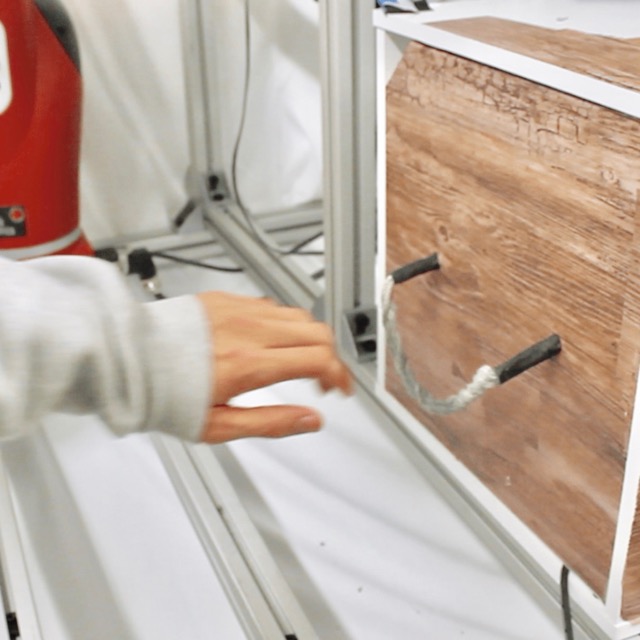}
  \includegraphics[origin=c,width=0.23\columnwidth]{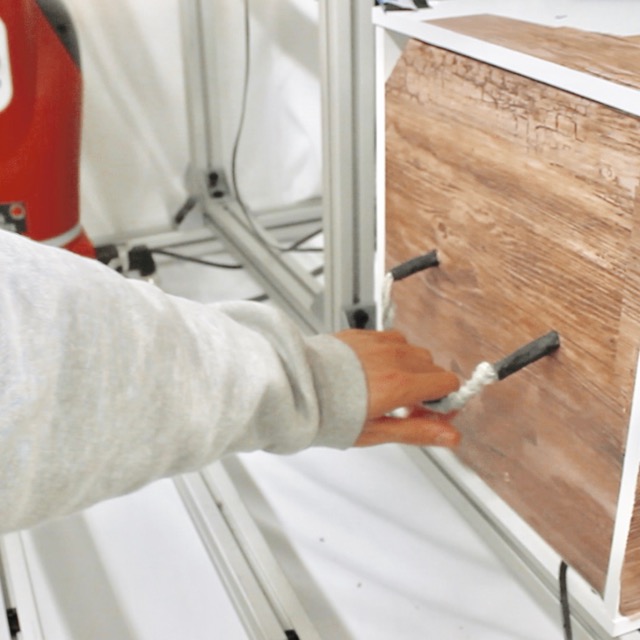}
  \includegraphics[origin=c,width=0.23\columnwidth]{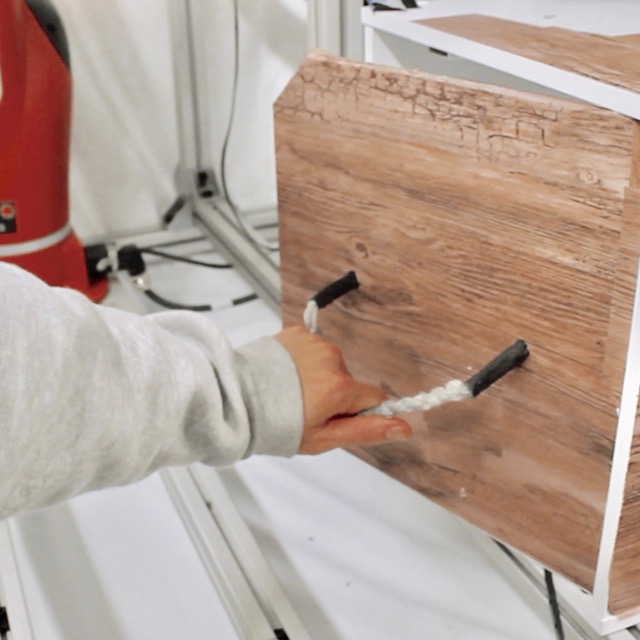}
  \caption{Opening door with flexible handle}
  \label{fig:door_opening}
\end{figure} 

The state space is all the joint angles of the hand, the Cartesian position of the arm, the current angle of the door, and last action taken. The action space is the position space of the hand and horizontal position of the wrist of the arm. The reward function is provided as 
\begin{gather*}
    r = -(d\theta)^2 - (x_{\text{arm}} - x_{\text{door}}) \\
    d\theta := \theta_{\text{door}} - \theta_{\text{closed}}
\end{gather*}

We define a trajectory as a success if at any point $d\theta > 30\degree$.

\subsection{Dynamixel Driven State Estimation}
For these tasks, we solve both the problem of state estimation and resetting using a setup with objects augmented with Dynamixel servo motors. These motors serve the dual purpose of resetting objects (such as the valve, box, or door) to their original positions and measuring the state of the object (angle of rotation or position).

\begin{figure}[!h]
  \centering
  \includegraphics[origin=c,width= 0.3\columnwidth]{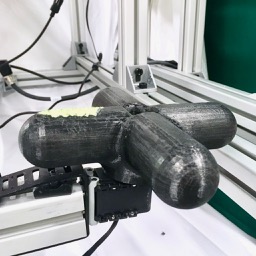}
  \includegraphics[origin=c,width= 0.3\columnwidth, ]{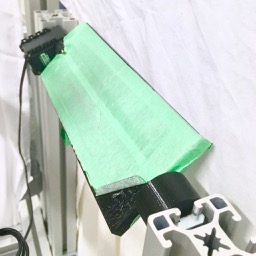} 
  \includegraphics[origin=c,width= 0.3\columnwidth, ]{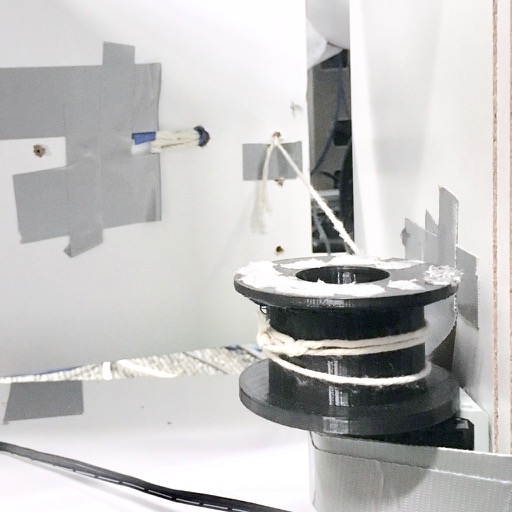} \\
  \caption{Dynamixel driven sensing and reset mechanisms. Left to Right: rigid valve, box, door.}
  \label{fig:reset_mechanisms}
\end{figure} 

\section{Algorithms}
\label{sec:algorithms}
In this work, we show that the tasks described in the previous section can be solved using model-free on-policy reinforcement learning algorithms. This requires 7 hours for valve turning, 4 for box flipping and 16 hours for door opening. We also demonstrate that the learning time can be significantly reduced by using a small number of human demonstrations.

\subsection{Policy Gradient}
\label{sec:NPG}
Reinforcement learning algorithms operate within the framework of Markov decision processes. An MDP is described using the tuple: \hbox{$\mathcal{M} = \lbrace \mathcal{S}, \mathcal{A}, \mathcal{R}, \mathcal{T}, \rho_0, \gamma \rbrace$}. Here, $\mathcal{S} \subseteq \bR^n$ and $\mathcal{A} \subseteq \bR^m$ represent the state and action spaces respectively. \hbox{$\mathcal{R}: \mathcal{S} \times \mathcal{A} \rightarrow \mathbb{R}$} is the reward function. $\mathcal{T}: \mathcal{S} \times \mathcal{A} \times \mathcal{S} \rightarrow \bR_{\geq 0}$ represents the stochastic unknown transition dynamics. The goal in reinforcement learning is to find a policy $\pi$ that describes how to choose actions in any given state, such that we maximize the sum of expected rewards. The performance of a policy is given by:
\begin{equation}
\eta(\pi) = \bE_{\tau \sim \cM^{\pi}} \Bigg[ \sum_{t=0}^\infty \gamma^t r_t \Bigg].
\end{equation}

In this work, we parameterize the policy using a neural network, and use a gradient ascent based approach to optimize (1). Simply performing gradient ascent on (1) is often referred to as vanilla policy gradient (REINFORCE)~\cite{williams92}, given by:

\begin{equation}
    \nabla_{\theta}\eta = \bE_{\cM^{\pith}} \Bigg[ \sum_{t=0}^T \nabla_{\theta}\log \pith(a_t|s_t) \sum_{t'=t}^T r(s_t, a_t) \Bigg]
\end{equation}

REINFORCE is known to be slow and ineffective. Natural policy gradient methods capture curvature information about the optimization landscape, thereby stabilizing the optimization process and enabling faster convergence. The natural policy gradient is computed by preconditioning the REINFORCE gradient with the inverse of the Fisher Information Matrix, which is defined as
\begin{equation}
    \mathcal{F}(\theta) = \mathbb{E}_{s,a}[\nabla_{\theta}\log \pith(a|s) \nabla_{\theta}\log \pith(a|s)^T].
\end{equation}

Thus, the natural policy gradient update rule is given by:
\begin{equation}
    \theta_{i+1} = \theta_{i} + \alpha \mathcal{F}^{-1}(\theta_i) \nabla_{\theta_i}\eta
\end{equation}
There are numerous variants of the natural policy gradient method ~\cite{trpo, peters2008natural}. For simplicity, we use the truncated natural policy gradient (TNPG) as described in ~\cite{simplicity}.

\subsection{Demonstration Augmented Policy Gradient}
\label{sec:dapg}
While the NPG algorithm is guaranteed to converge asymptotically to at least a locally optimal policy, the rate of convergence could be very slow. In particular, each update step in NPG requires interacting with the environment to collect on-policy data which may be slow for some tasks. In such cases, we would like to accelerate the learning process by using various forms of prior knowledge.

One way to incorporate human priors is through the use of demonstration data, obtained through say kinesthetic teaching. In prior work, we developed the demonstration augmented policy gradient (DAPG) algorithm which combines reinforcement learning with imitation learning using the demonstration data. Let $\cD = \lbrace{ (s_t^i, a_t^i, r_t^i) \rbrace}$ denote the demonstration dataset. Let $\cD^\pi$ denote the on-policy dataset obtained by rolling out $\pi$. DAPG starts by pre-training the policy $\pi$ with behavior cloning on this demonstration data $\cD$. This pre-trained policy $\pi$ is subsequently finetuned using an augmented policy gradient. 

DAPG first constructs an augmented ``vanilla'' gradient as:
\begin{equation}
    \begin{aligned}
        g_{aug} = & \sum_{(s,a) \in \cD^\pi} \nabla_\theta \ln \pith(a|s) A^\pi(s,a) + \\
        & \sum_{(s,a) \in \cD} \nabla_\theta \ln \pith(a|s) w(s,a).
    \end{aligned}
\end{equation}
We choose $w(s,a) = \lambda_0 \lambda_1^k \max_{(s',a')\in \cD^\pi}A^\pi(s',a')$, where $\lambda_0$ and $\lambda_1$ are hyperparameters, and $k$ is the iteration counter. Subsequently, the policy is updated by preconditioning this augmented gradient with the Fisher information matrix as described in Section~\ref{sec:NPG}

The second term in $g_{aug}$ encourages the policy to be close to the actions taken by experts on states visited by the experts, throughout the learning process. Thus, it can be interpreted as reward shaping with a shaping similar to a trajectory tracking cost. 
This DAPG algorithm has been shown to significantly accelerate the learning by improving exploration, and exceed the performance of the demonstrations while still retaining their stylistic aspects -- all in simulation. In this work, we demonstrate that this algorithm provides a practical and useful way of accelerating deep RL on real hardware to solve challenging manipulation problems.  

\section{Experimental Results and Analysis}

The goal of our experiments is to empirically address the following research questions:
\begin{itemize}
    \item Can model-free deep RL provide a practical means for learning various dexterous manipulation behaviors directly in the real world?
    \item Can model-free deep RL algorithms learn on different hardware platforms and on different physical setups?
    \item Can we accelerate the learning process using a small number of demonstrations obtained through kinesthetic teaching?
    \item How do particular design choices of the reward function and actuation space affect learning? 
\end{itemize}
To do so, we utilize the tasks in Section~\ref{sec:tasks} and algorithms described in Section~\ref{sec:algorithms}. Additional details can be found at the supplementary website \url{https://sites.google.com/view/deeprl-handmanipulation}

\subsection{Model-Free Deep RL}

\begin{figure}[!h]
  \centering
    \includegraphics[width=1\columnwidth]{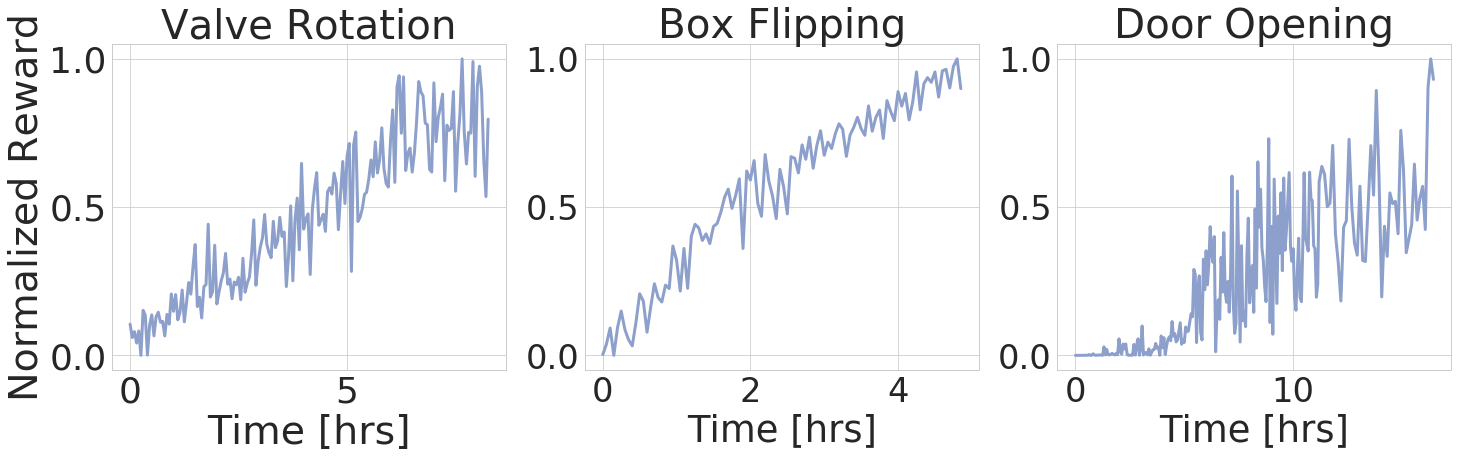}
  \caption{Learning progress with model-free RL from scratch using the NPG algorithm. The policies reach 100\% average success rates after 7.4 hours for valve rotation and 4.1 hours for box flipping and 15.6 hours for door opening.}
  \label{fig:learningcurves_dclaw_tnpg}
\end{figure}

First, we explore the performance of model-free deep reinforcement learning on our suite of hardware tasks. The learning progress is depicted in Fig~\ref{fig:learningcurves_dclaw_tnpg} and also in the accompanying video. We find that, somewhat surprisingly, model-free deep RL can acquire coherent manipulation skills in the time scales of a few hours (7 hours for turning a valve, 4 hours to flip a box, 16 hours for opening a door). These training times are evaluated once the deterministic policy achieves 100\% success rate over 10 evaluation rollouts, according to the success metrics defined in Section~\ref{sec:tasks} (Fig~\ref{fig:Train_times}). The algorithm is robust and did not require extensive hyperparameter optimization. The only hyperparameter that was tuned was the initial variance of the policy for exploration. We analyze specific design choices in the reward function and actuation scheme in Section~\ref{sec:analysis}.

\begin{figure}[!h]
  \centering
  \includegraphics[origin=c,width=0.23\columnwidth]{media/dclaw_scratch_screw_filmstrip/dclaw_screwscratch1.jpg}
  \includegraphics[origin=c,width=0.23\columnwidth]{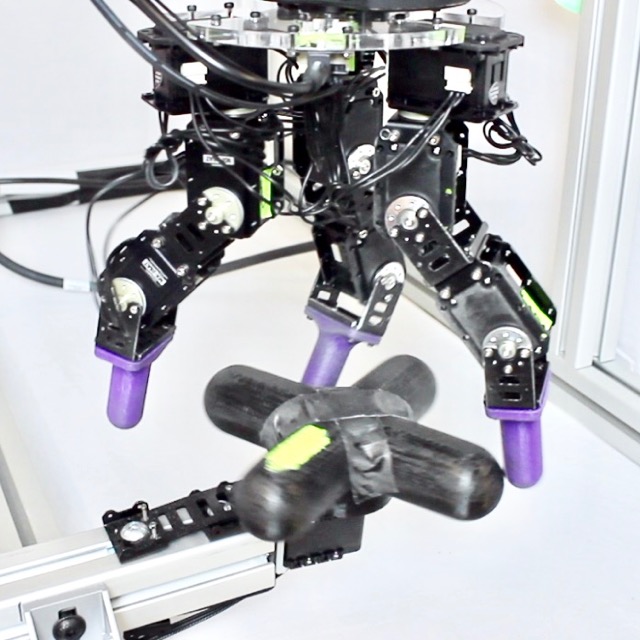}
  \includegraphics[origin=c,width=0.23\columnwidth]{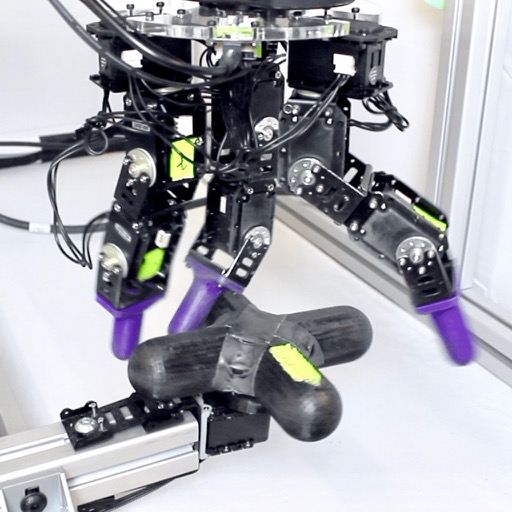}
  \includegraphics[origin=c,width=0.23\columnwidth]{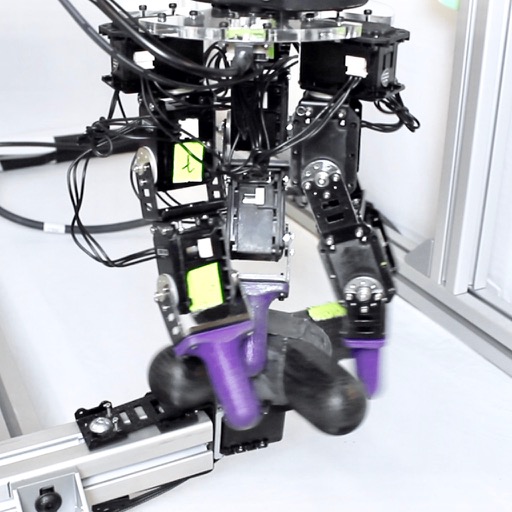}
  \caption{Visualization of Dclaw policy that has learned to turn a valve after 7 hours of training. The fingers learn to alternately move in and out to turn the valve.}
  \label{fig:screwing_dclaw}
\end{figure}

\begin{figure}[!h]
  \centering
  \includegraphics[origin=c,width=0.23\columnwidth]{media/flipping_film_strip/flip_film1.jpg}
  \includegraphics[origin=c,width=0.23\columnwidth]{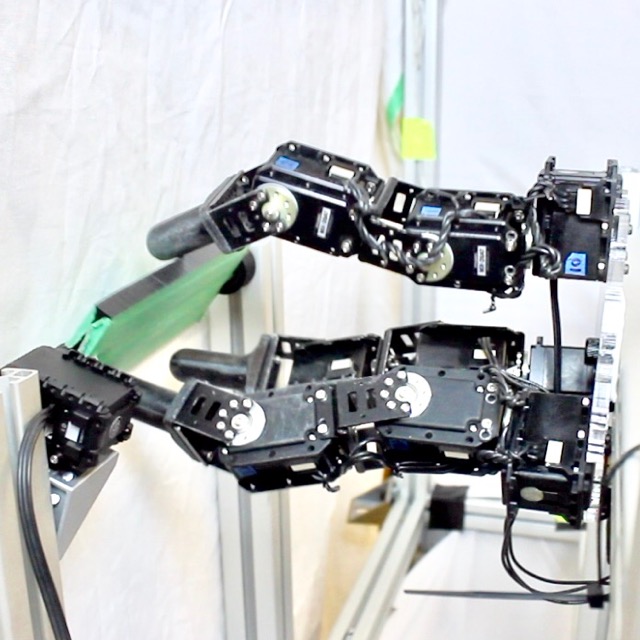}
  \includegraphics[origin=c,width=0.23\columnwidth]{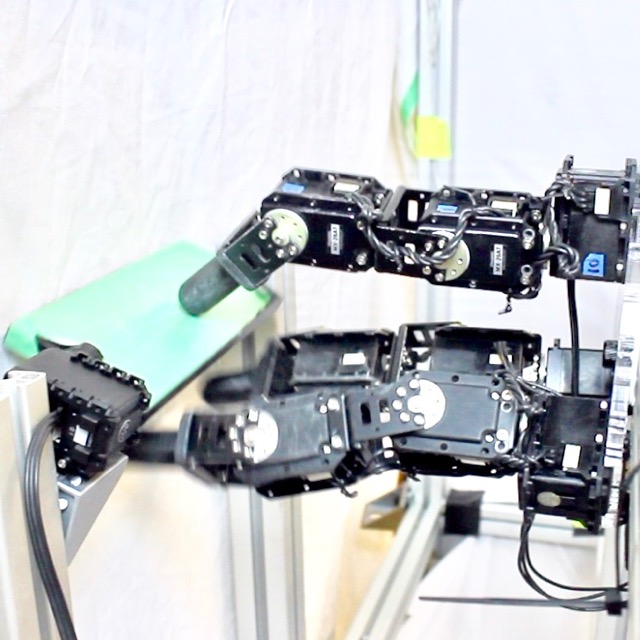}
  \includegraphics[origin=c,width=0.23\columnwidth]{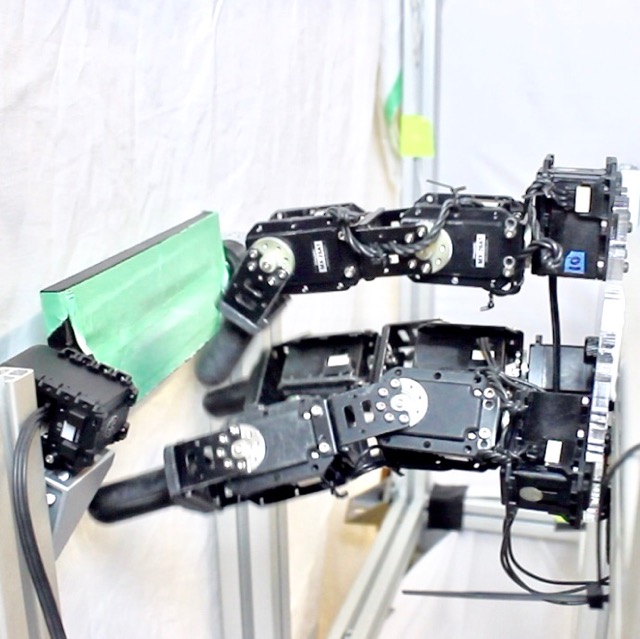}
  \caption{Visualization of Dclaw that has learned to flip a box after 4 hours of training. The Dclaw learns to extend its bottom two fingers and push its top finger forwards,  then lower its bottom two fingers while pushing downwards with its top finger.}
  \label{fig:flipping_dclaw}
\end{figure}

\begin{figure}[!h]
  \centering
    \includegraphics[origin=c,width=0.23\columnwidth]{media/door_film_strip/door_film1.jpg}
  \includegraphics[origin=c,width=0.23\columnwidth]{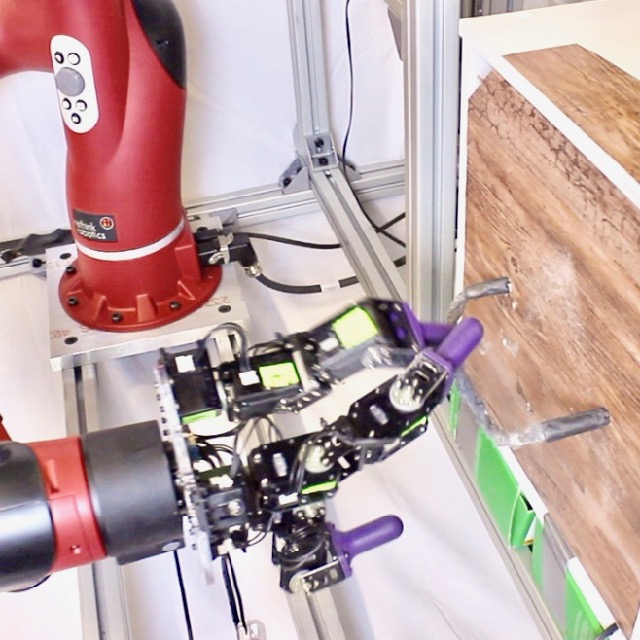}
  \includegraphics[origin=c,width=0.23\columnwidth]{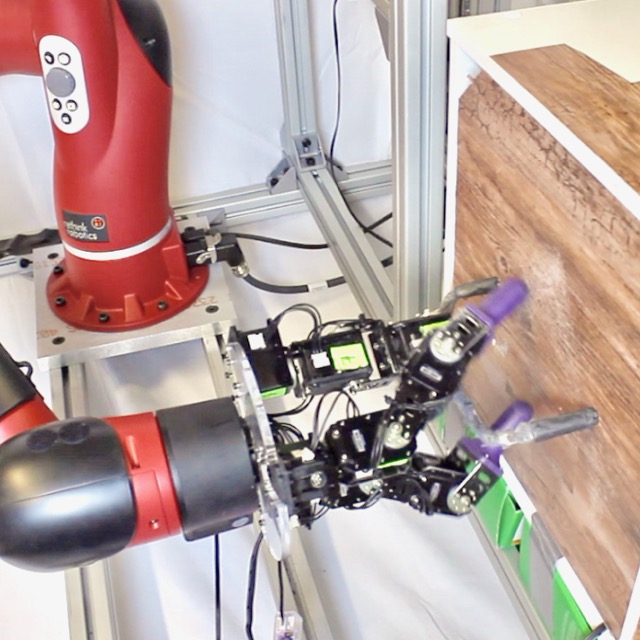}
  \includegraphics[origin=c,width=0.23\columnwidth]{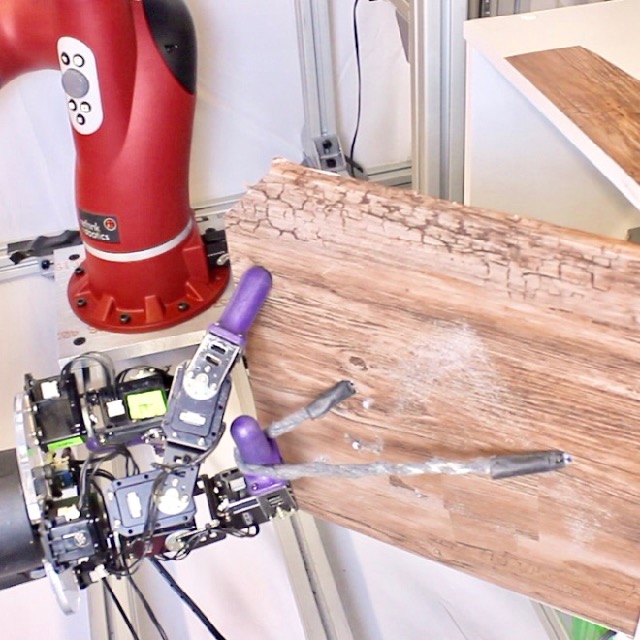}
  \caption{Visualization of a policy that has learned to open a door after 16 hours of training. The robot learns to move towards the door, grasp the deformable handle, and then pull the door open.}
  \label{fig:door_dclaw}
\end{figure}

We find that for the valve and the box flipping, the learning is able to monotonically improve on the continuous reward signal, whereas for the door the learning is more challenging, given that reward is only obtained when the door is actually opened. The agent has to consistently pull open the door in order to see the reward, leading to the large spikes in learning as seen in Fig.~\ref{fig:learningcurves_dclaw_tnpg}.

The learned finger gaits that we observe do have interesting coordination patterns. The tasks require that the fingers move quickly and in a coordinated way so as to rotate the objects and then move out of the way. This behavior can be appreciated in the accompanying video on the supplementary website.

\subsection{Learning on Different Hardware and Different Materials}

\begin{figure}[!h]
  \centering
  \includegraphics[origin=c,width=0.49\columnwidth]{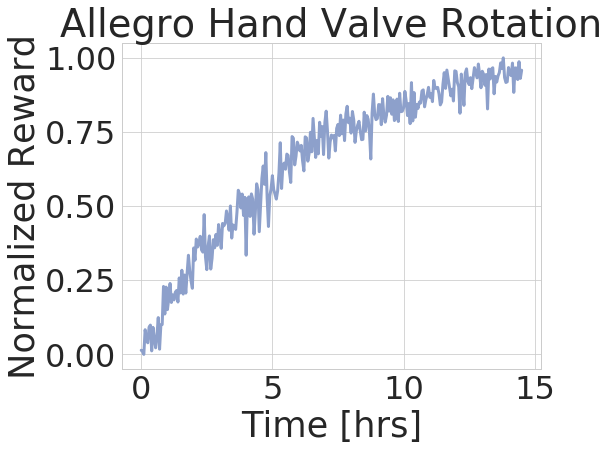}
  \includegraphics[origin=c,width=0.47\columnwidth]{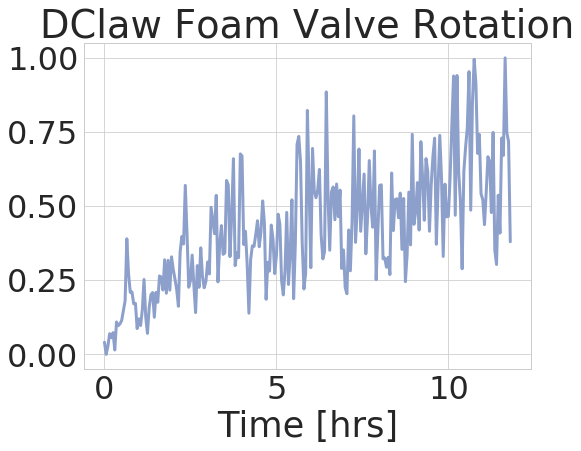}
  \caption{Left: learning progress of Allegro hand rotating rigid screw. Right: learning progress of Dclaw rotating the foam valve. The rewards have been normalized such that a random policy achieves score of 0 and the final trained policy achieves a score of $1$.}
  \label{fig:Allegro_rigid_screw}
\end{figure}

To illustrate the ability of model-free RL algorithms to be applied easily to new scenarios and robots, we evaluated the valve task using the exact same deep RL algorithm with a different hand -- the 4-fingered Allegro hand. This hand has 16 DoFs and is also anthropomorphic. We find that the Allegro hand was able to learn this task, as illustrated in Fig.~\ref{fig:Allegro_rigid_screw}, in comparable time as the Dclaw.

\begin{figure}[h]
  \centering
  \includegraphics[origin=c,width=0.24\columnwidth]{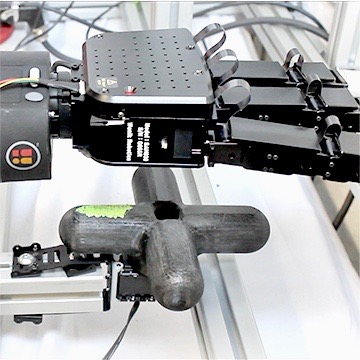}
  \includegraphics[origin=c,width=0.24\columnwidth]{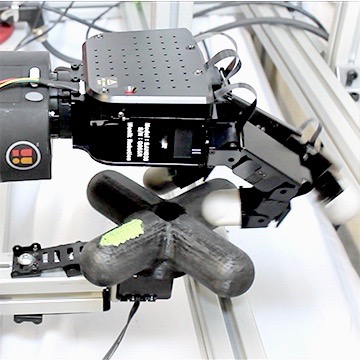}
  \includegraphics[origin=c,width=0.24\columnwidth]{media/allegro_screwing_filmstrip/allegro_rValve_film3.jpg}
  \includegraphics[origin=c,width=0.24\columnwidth]{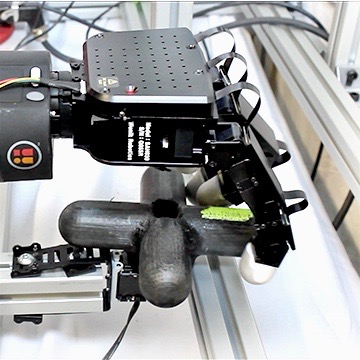}
  \caption{The Allegro hand learns to rotate a valve 180\degree.}
  \label{fig:Allegro_rigid_screw_film}
\end{figure}

Both systems are able to quickly learn the right behavior with the same hyperparameters. While morphology does indeed have an effect on rate of learning, we find that it is not a hindrance to eventually learning the task. The easy adaptability of these algorithms is extremely important, since we don't need to construct an accurate simulation or model of each new robot, and can simply run the same algorithm repeatedly to learn new behaviors. 

\begin{figure}[!h]
  \centering
  \includegraphics[origin=c,width=0.24\columnwidth]{media/foam_screw_filmstrip/foam_film1.jpg}
  \includegraphics[origin=c,width=0.24\columnwidth]{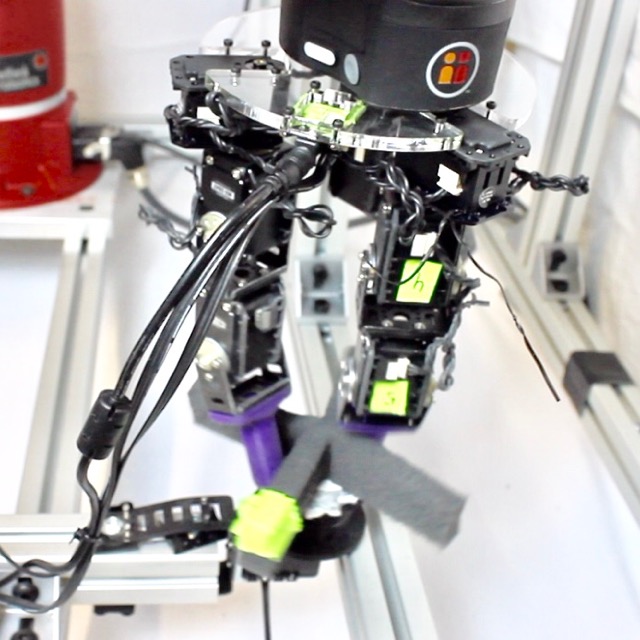}
  \includegraphics[origin=c,width=0.24\columnwidth]{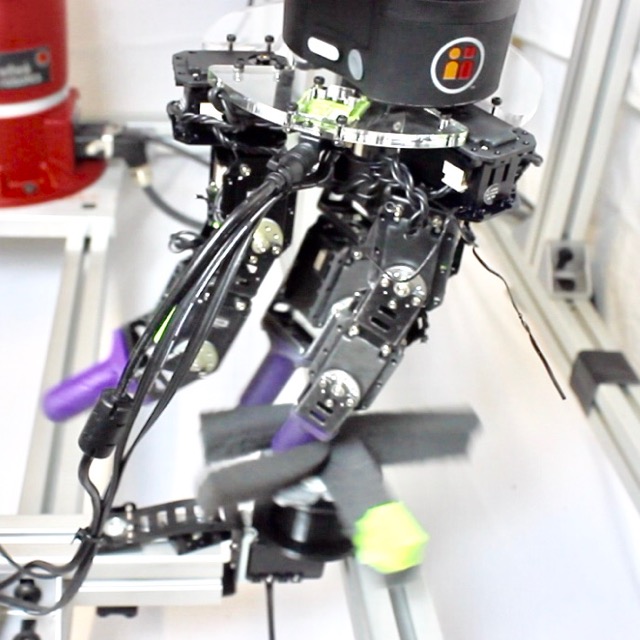}
  \includegraphics[origin=c,width=0.24\columnwidth]{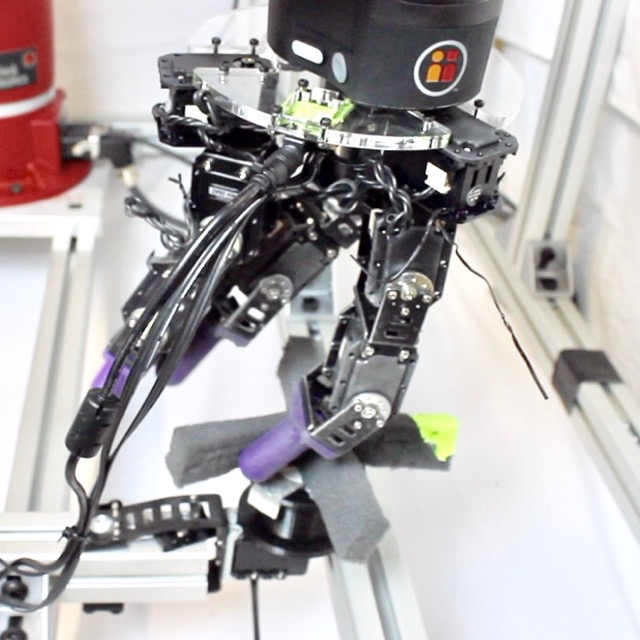}
  \caption{DClaw learns to rotate a foam valve despite its deformable structure. The claw learns to focus its manipulation on the center of the valve where there is more rigidity.}
  \label{fig:Allegro_foam_screw}
\end{figure}

Besides changing the robot's morphology, we can also modify the object that is manipulated. We evaluate whether model-free RL algorithms can be effective at learning with a different valve material, such as soft foam (Fig.~\ref{fig:Allegro_foam_screw}). The contact dynamics with such a deformable material are hard to simulate and the hand can deform the valve in many different directions, making the actual manipulation task challenging. We see that model-free RL is able to learn manipulation with the foam valve  effectively, even generating behaviors that exploit the deformable structure of the object. 

\subsection{Accelerating Learning with Demonstrations}

While we find that model-free deep RL is generally practical in the real world, the number of samples can be further reduced by employing demonstrations, as discussed in Section~\ref{sec:dapg}. In order to understand the role of demonstrations, we collected 20 demonstrations for each task via kinesthetic teaching. 

\begin{figure}[!h]
  \centering
  \includegraphics[origin=c,width=0.4\columnwidth]{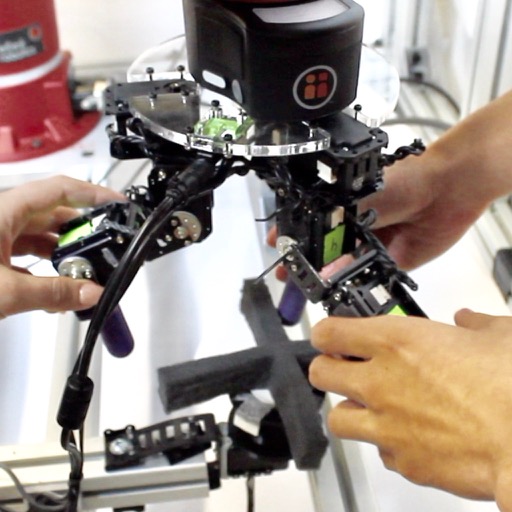}
  \caption{Kinesthetic teaching was used to obtain demonstrations with the Dclaw.}
\end{figure} 

\begin{figure}[!h]
  \centering
  \includegraphics[origin=c,width= 0.49\columnwidth]{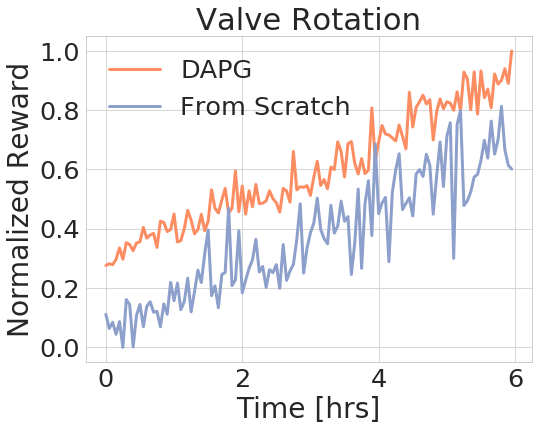}
  \includegraphics[origin=c,width= 0.46\columnwidth]{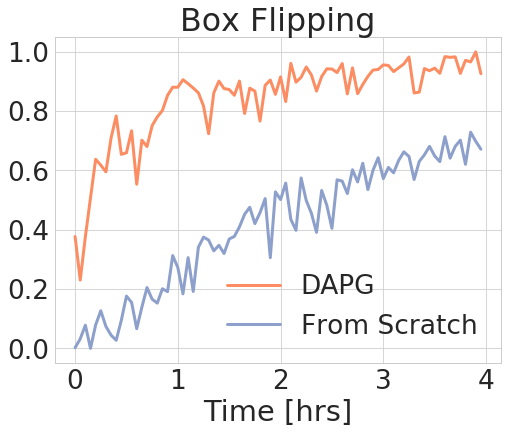}
  \caption{Learning progress with training from scratch using NPG and using DAPG. The performances have been normalized such that a random policy achieves a score of $0$ and the best DAPG policy gets a score of $1$.}
  \label{fig:DAPG_results}
\end{figure}

These demonstrations are slow and suboptimal, but can still provide guidance for exploration and help guide the learning process. With only a few demonstrations, we see that demonstration augmented policy gradient (DAPG) can speed up learning significantly, dropping learning time by 2x (Fig~\ref{fig:DAPG_results}). We record training times across tasks using the previously defined success metrics in (Fig~\ref{fig:Train_times}).

\begin{figure}[!h]
\caption{Training Times Across Tasks [hrs]. Training time is determined using the success metrics defined in Section IV. A training run is complete once the deterministic policy achieves 100\% success rate over 10 evaluation rollouts.} 
\centering 
\begin{tabular}{c c c} 
\hline\hline 
Task & From Scratch & DAPG\\ [0.5ex] 
\hline 
Valve & 7.4 & 3.0 \\ 
Box & 4.1 & 1.5 \\
Door & 15.6 & -----\\
[1ex] 
\hline 
\end{tabular}
\label{fig:Train_times} 
\end{figure}

The efficiency of DAPG comes from the fact that the behavior cloning initialization gives the agent a rough idea of how to solve the task, and the augmented loss function guides learning through several iterations. The behaviors learned are also more gentle and legible to humans than behaviors learned via training from scratch, which is clearly displayed in the accompanying video.

\begin{figure}[!h]
  \centering
  \includegraphics[origin=c,width=0.49\columnwidth]{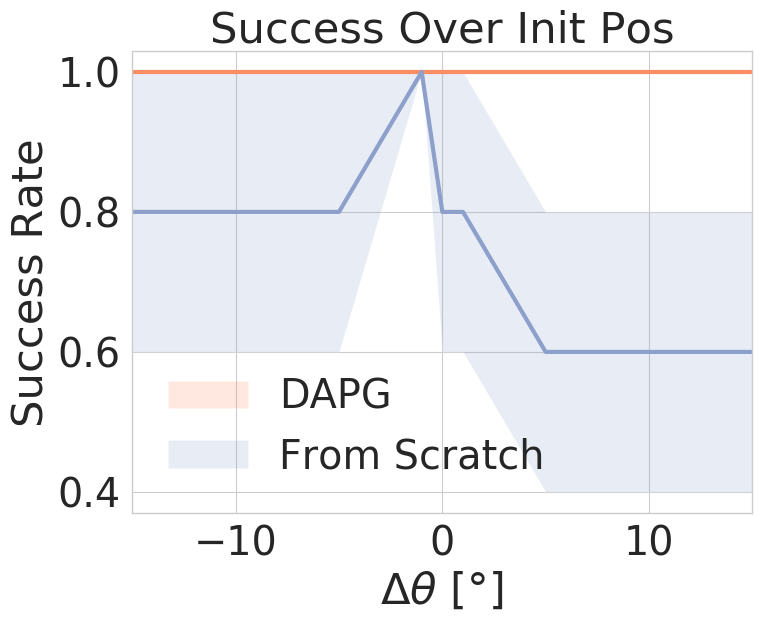}
   \includegraphics[origin=c,width=0.49\columnwidth]{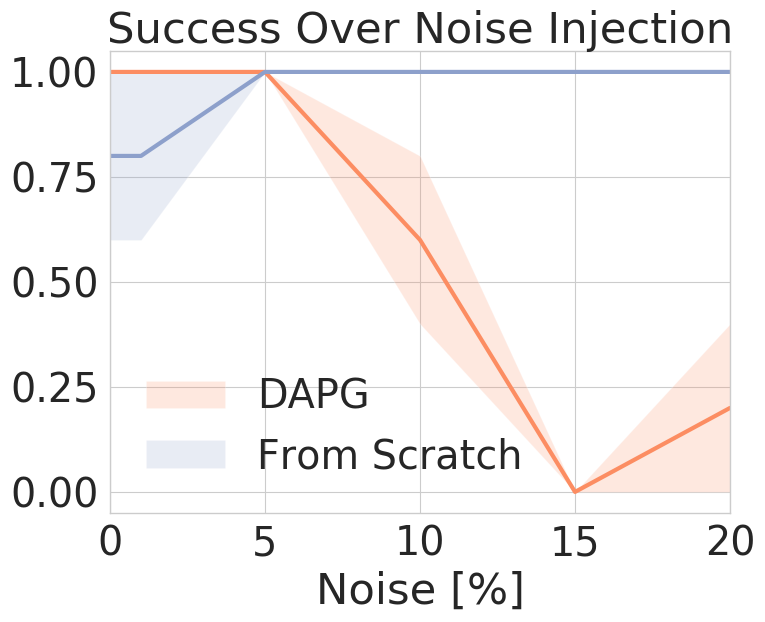}
  \caption{Plots showing robustness of DAPG vs learning from scratch for the valve rotation task. Left: Variation of success with change in valve initial position (degrees). Right: Variation of success with x\% uniformly random noise injected into the observation and action space. DAPG is more robust with change in initial valve position, but is less robust as we add more noise.}
  \label{fig:robustness}
\end{figure}

To better understand the learned behaviors, we also evaluated the robustness of these behaviors to variations in the initial position of the valve, and to noise injected into actions and observations (Fig~\ref{fig:robustness}).

\begin{figure}[!h]
  \centering
\includegraphics[width=0.8\columnwidth]{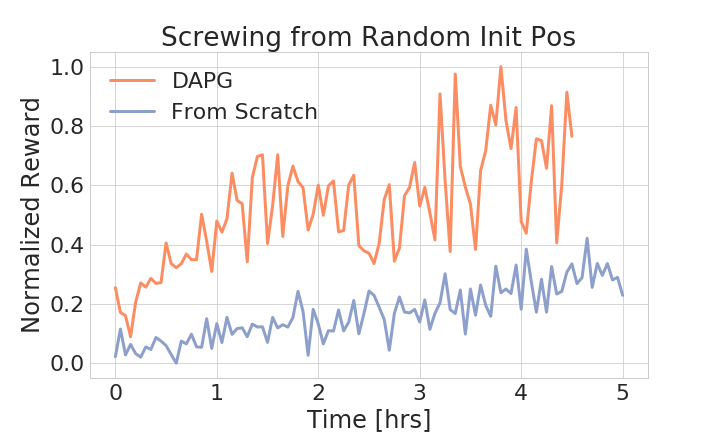} \\
    \caption{Training DClaw to turn a valve from a single randomly sampled initial position between [-45\degree, 45\degree] to 180\degree. DAPG was trained with 20 demos that turned the valve from initial positions sampled uniformly at random between [-45\degree, 45\degree] to 180\degree.}
  \label{fig:DAPG_generalization_results}
\end{figure}

We also considered collecting demonstrations from a wider range of initial configurations of the environment, and using this wider demo set for DAPG. The demonstrations were collected with initial valve positions in the range $[-\frac{\pi}{4}, \frac{\pi}{4}]$. This paradigm also works well (Fig~\ref{fig:DAPG_generalization_results}). Unsurprisingly, it is not as effective as using a number of demonstrations in the same environment configuration, but is still able to learn well. (Fig~\ref{fig:DAPG_results}).

\subsection{Performance with Simulated Training}
\label{sec:sim2real}

While the main goal of this work is to study how model-free RL can be used to learn complex policies directly on real hardware, we also evaluated training in simulation and transfer, employing randomization to allow for transfer~\cite{epopt,cad2rl}. This requires modeling the task in a simulator and manually choosing the parameters to randomize.

\begin{figure}[!h]
  \centering
\includegraphics[width=0.8\columnwidth]{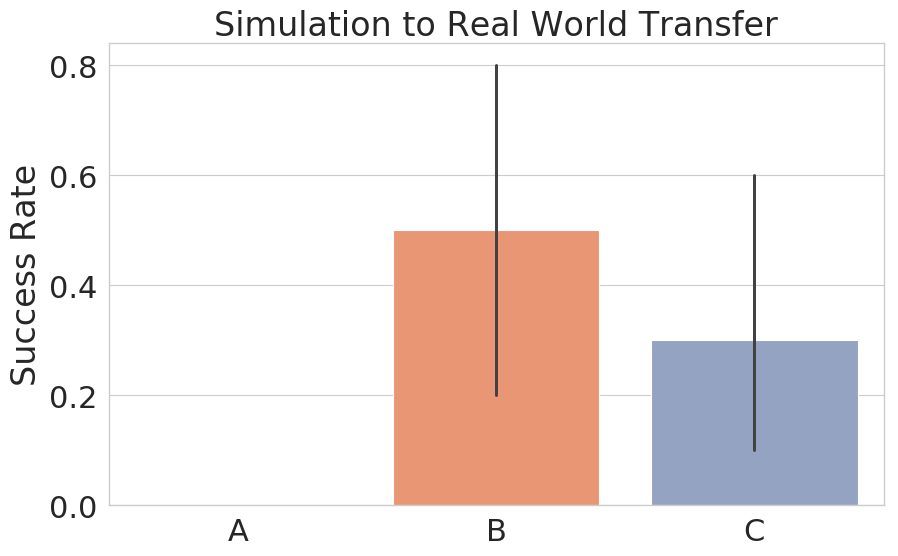} \\
  \caption{Success rates using different sim2real transfer strategies for valve turning with Dclaw. A: No domain randomization. B: Randomization of position control PID parameters and friction. C: Same as B, but also including the previous action as part of the state space.}
  \label{fig:sim2real_analysis}
\end{figure} 

We see in Fig.~\ref{fig:sim2real_analysis} that the randomization of PID parameters and friction is crucial for effective transfer.

While simulation to real transfer enabled by randomization is an appealing option, especially for fragile robots, it has a number of limitations. First, the resulting policies can end up being overly conservative due to the randomization, a phenomenon that has been widely observed in the field of robust control. Second, the particular choice of parameters to randomize is crucial for good results, and insights from one task or problem domain may not transfer to others. Third, increasing the amount of randomization results in more complex models tremendously increasing the training time and required computational resources, as discussed in Section~\ref{sec:related}. Directly training in the real world may be more efficient and lead to better policies. Finally, and perhaps most importantly, an accurate simulator must be constructed manually, with each new task modeled by hand in the simulation, which requires substantial time and expertise. For tasks such as valve rotation with the foam valve or door opening with a soft handle, creating the simulation itself is very challenging.

\subsection{Design Choices}
\label{sec:analysis}
To understand the design choices needed to effectively train dexterous hand manipulation systems with model-free RL, we analyzed different factors which contribute to learning. We performed this analysis in simulation in order to choose the right schemes for real world training. 

\subsubsection{Choices of Action Space}
The choice of actuation space often makes an impact on learning progress . For hand manipulation it also greatly affects the smoothness, and hence sustainability, of the hardware. In our results, we end up using position control since it induces the fewest vibrations and is easiest to learn with. 

In order to better understand the rationale behind this, we consider a comparison between using controlling position and torque controllers as well as their higher order derivatives. We compare the vibrations induced by each of these control schemes by measuring the sum of the magnitudes of the highest Fourier coefficients of sample trajectories (joint angles) induced by random trajectories. 

\begin{figure}[!h]
  \centering
  \includegraphics[origin=c,width=0.49\columnwidth]{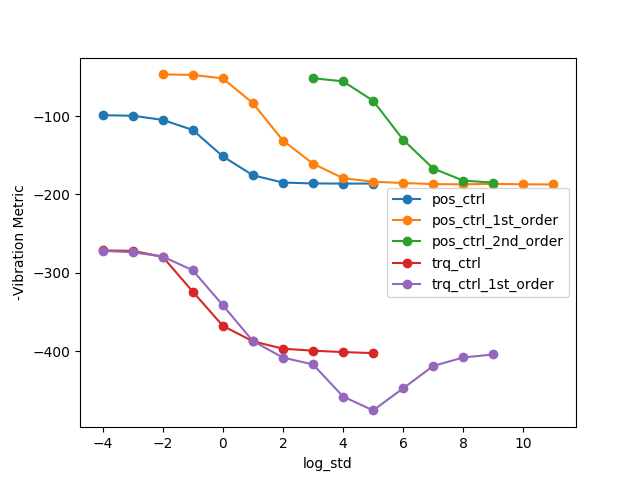}
  \includegraphics[origin=c,width=0.49\columnwidth]{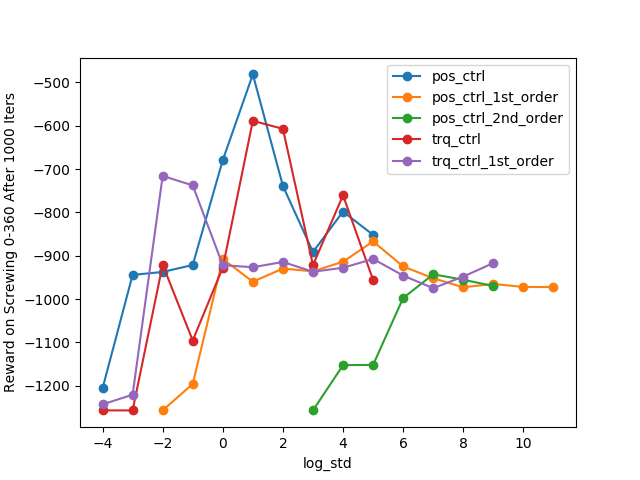}
  \caption{Left: Analysis of vibrations induced by different actuation schemes in simulation. Higher metric indicates lower vibrations. We find that position control is able to induce significantly lower vibrations than torque control, making it safer to run on hardware. Right: Analysis of rewards attained in simulation after training using the control scheme on Dclaw valve rotating task.}
  \label{fig:actuation_analysis}
\end{figure} 

We see that position control has the lowest vibration amongst all the choices of control schemes, and is also able to achieve significantly better performance. This is likely because we are using a stabilizing PID control at the low level to do position control which reduces the load on the learning algorithm. We also see that it is harder to learn a policy when controlling higher order derivatives, and that it is easier to learn with position control than with torque control.

\subsubsection{Impact of Reward Function}
We also investigated the effect of the reward function on learning progress. To provide some intuition about the different choice of reward functions, we show a comparison between 3 different reward functions for learning. We evaluate learning progress for these three types of reward functions in simulation, as a means for choosing an appropriate form of reward for real world training. 

\begin{enumerate}
    \item $r_1 = -\|\theta - \theta_{goal}\|_2$
    \item  $r_2 = r_1 + 10*\mathbb{1}_{\{r_1 < 0.1\}} + 50*\mathbb{1}_{\{r_1 < 0.05\}}$ 
    \item  $r_3 = r_2 - \|v\|_2$
\end{enumerate}

\begin{figure}[!h]
  \centering
  \includegraphics[origin=c,width=0.7\columnwidth]{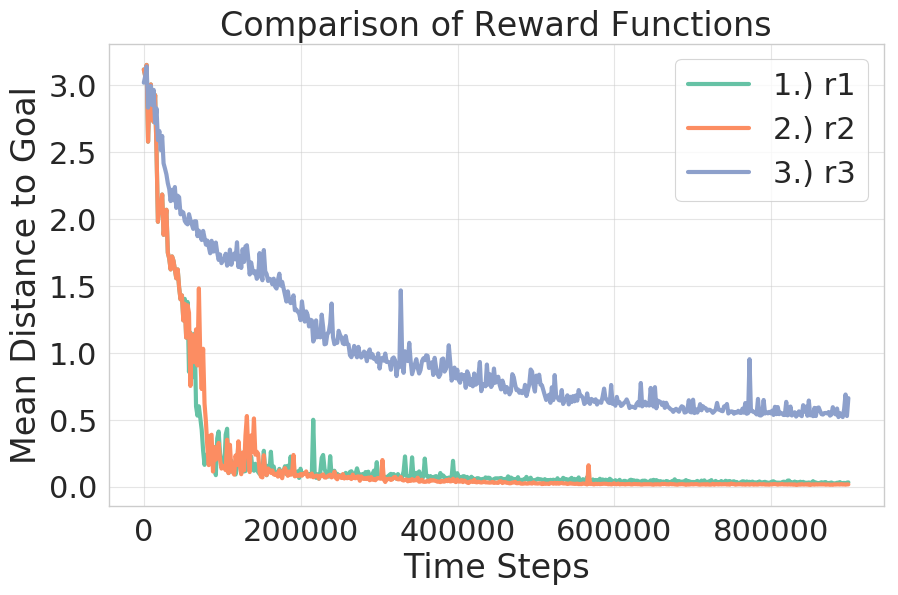}
  \caption{Analysis of learning progress in simulation with different reward functions.}
  \label{fig:reward_analysis}
\end{figure} 

We find that learning is most effective with using either option 1 or 2, and is slower with a control cost. The control cost ensures smoother operation, but at the cost of efficiency in learning, since optimization of control penalties results in reduction of exploration. In real world experiments we found that training without a control cost still produced safe behaviors.
 
\section{Discussion and Future Work}

In this work, we study real-world model-free reinforcement learning as a means to learn complex dexterous hand manipulation behaviors. We show that model-free deep RL algorithms can provide a practical, efficient, and general method for learning with high dimensional multi-fingered hands. Model-free RL algorithms can easily be applied to a variety of low-cost hands, and solve challenging tasks that are hard to simulate accurately. This kind of generality and minimal manual engineering may be a key ingredient in endowing robots with the kinds of large skill repertoires they need to be useful in open-world environments such as homes, offices, and hospitals. We also show that the sample complexity of model-free RL can be substantially reduced with suboptimal kinesthetic demonstrations, while also improving the resulting motion quality.

There are several exciting avenues for future work. Firstly, deep neural networks can not only represent complex skills, but can also high-dimensional inputs such as images. Studying dexterous manipulation with visual observations is an exciting direction for future work. Furthermore, while we learn each skill individually in our experiments, an exciting future direction is to pool experience across multiple skills and study the ability to acquire new behaviors more efficiently in a multi-task setting, which is an important stepping stone toward generalist robots endowed with large behavioral repertoires.

\bibliographystyle{ieeetr}
\bibliography{references}

\end{document}


\appendix
\section*{Appendix A. Network Architecture and Training}
For encoders $\text{Enc}_1$ and $\text{Enc}_2$ in simulation we use stride-2 convolutions with a $5\times5$ kernel. We perform 4 convolutions with filter sizes 64, 128, 256, and 512 followed by two fully-connected layers of size 1024. We use LeakyReLU activations with leak 0.2 for all layers. The translation module $T(z_1, z_2)$ consists of one hidden layer of size 1024 with input as the concatenation of $z_1$ and $z_2$ and output of size 1024. For the decoder $\text{Dec}$ in simulation we have a fully connected layer from the input to four fractionally-strided convolutions with filter sizes 256, 128, 64, 3 and stride $\frac{1}{2}$. We have skip connections from every layer in the context encoder $\text{Enc}_2$ to its corresponding layer in the decoder $\text{Dec}$ by concatenation along the filter dimension. 

For real world images, the encoders perform 4 convolutions with filter sizes 32, 16, 16, 8 and strides 1, 2, 1, 2 respectively. All fully connected layers and feature layers are size 100 instead of 1024. The decoder uses fractionally-strided convolutions with filter sizes 16, 16, 32, 3 with strides $\frac{1}{2}$, 1, $\frac{1}{2}$, 1 respectively. For the real world model only, we apply dropout for every fully connected layer with keep probability 0.5, and we tie the weights of $\text{Enc}_1$ and $\text{Enc}_2$. 

We train using the ADAM optimizer with learning rate $10^{-4}$. We train using 3000 videos for reach, 4500 videos for simulated push, 894 videos for sweep, 180 videos for simulated push with real videos, and 135 videos for real push with real videos.

\section*{Appendix B. Ablation Study}
To evaluate that the different loss functions while training our translation model, and the different components for the reward function while performing imitation, we performed ablations by removing these components one by one during model training or policy learning. To understand the importance of the translation cost, we remove cost $\mathcal{L}_{\text{trans}}$, to understand whether features $z_3$ need to be properly aligned we remove model losses $\mathcal{L}_{\text{rec}}$ and $\mathcal{L}_{\text{align}}$. We see that the removal of each of these losses significantly hurts the performance of subsequent imitation. On removing the feature tracking loss $\hat{R}_\text{feat}$ or the image tracking loss $\hat{R}_\text{image}$ we see that overall performance across tasks is worse.

\begin{figure}[h!]
\centering
  \includegraphics[width=0.95\textwidth]{plot/barablation.png} 
 
  \caption{Ablations on model losses and reward functions for the reaching, pushing and pushing with real world demonstrations tasks. Across tasks, all components of the model are necessary for success.}
  \label{fig:allablations}
\end{figure}

\newpage
\section*{Appendix C. Sample Videos}

\subsection*{Reach Simulation}
\begin{figure}[h!]
  \centering
  \includegraphics[width=0.9\textwidth]{diagrams/reach/reach6.jpg}
  \includegraphics[width=0.9\textwidth]{diagrams/reach/reach8.jpg}
  \caption{Example expert training demonstrations from different viewpoints with variations in color, distractor objects, and goal position.}
\end{figure}

\begin{figure}[h!]
  \small{Source Video} \hspace*{0.73cm}
  \vcenteredinclude{
  \includegraphics[width=0.8\textwidth]{diagrams/reach/reachsrc16.jpg}}\\
  \small{Target Context $o_0$} \hspace*{0.45cm}
  \vcenteredinclude{\includegraphics[width=0.145\textwidth]{diagrams/reach/reachctx16.png}}\\
  \small{Translated Video} \hspace*{0.3cm}
  \vcenteredinclude{\includegraphics[width=0.8\textwidth]{diagrams/reach/reachtrans16.jpg}}\\
  
  \small{Source Video} \hspace*{0.73cm}
  \vcenteredinclude{
  \includegraphics[width=0.8\textwidth]{diagrams/reach/reachsrc21.jpg}}\\
  \small{Target Context $o_0$} \hspace*{0.45cm}
  \vcenteredinclude{\includegraphics[width=0.145\textwidth]{diagrams/reach/reachctx21.png}}\\
  \small{Translated Video} \hspace*{0.3cm}
  \vcenteredinclude{\includegraphics[width=0.8\textwidth]{diagrams/reach/reachtrans21.jpg}}\\
  \caption{Example illustrations of demonstrations for a reaching task (top) being performed in a new context (middle), with the translated observation sequences (bottom).}
\end{figure}
\newpage

\subsection*{Push Simulation}
\begin{figure}[h!]
  \centering
  \includegraphics[width=0.9\textwidth]{diagrams/push/push0.jpg}
  \includegraphics[width=0.9\textwidth]{diagrams/push/push6.jpg}
  \caption{Example expert training demonstrations from different viewpoints with variations in distractor objects, start and goal position.}
\end{figure}

\begin{figure}[h!]
  \small{Source Video} \hspace*{0.73cm}
  \vcenteredinclude{
  \includegraphics[width=0.8\textwidth]{diagrams/push/push2src.jpg}}\\
  \small{Target Context $o_0$} \hspace*{0.45cm}
  \vcenteredinclude{\includegraphics[width=0.145\textwidth]{diagrams/push/push2ctx.png}}\\
  \small{Translated Video} \hspace*{0.3cm}
  \vcenteredinclude{\includegraphics[width=0.8\textwidth]{diagrams/push/push2trans.jpg}}\\
  
  \small{Source Video} \hspace*{0.73cm}
  \vcenteredinclude{
  \includegraphics[width=0.8\textwidth]{diagrams/push/push6src.jpg}}\\
  \small{Target Context $o_0$} \hspace*{0.45cm}
  \vcenteredinclude{\includegraphics[width=0.145\textwidth]{diagrams/push/push6ctx.png}}\\
  \small{Translated Video} \hspace*{0.3cm}
  \vcenteredinclude{\includegraphics[width=0.8\textwidth]{diagrams/push/push6trans.jpg}}\\
  \caption{Example illustrations of demonstrations for a pushing task (top) being performed in a new context (middle), with the translated observation sequences (bottom).}
\end{figure}

\newpage
\subsection*{Sweep Simulation}
\begin{figure}[h!]
  \centering
  \includegraphics[width=0.9\textwidth]{diagrams/sweep/sweep5.jpg}
  \includegraphics[width=0.9\textwidth]{diagrams/sweep/sweep7.jpg}
  \caption{Example expert training demonstrations from different viewpoints.}
\end{figure}

\begin{figure}[h]
  \small{Source Video} \hspace*{0.73cm}
  \vcenteredinclude{
  \includegraphics[width=0.8\textwidth]{diagrams/sweep/sweep25src.jpg}}\\
  \small{Target Context $o_0$} \hspace*{0.45cm}
  \vcenteredinclude{\includegraphics[width=0.145\textwidth]{diagrams/sweep/sweep25ctx.png}}\\
  \small{Translated Video} \hspace*{0.3cm}
  \vcenteredinclude{\includegraphics[width=0.8\textwidth]{diagrams/sweep/sweep25trans.jpg}}\\
  
  \small{Source Video} \hspace*{0.73cm}
  \vcenteredinclude{
  \includegraphics[width=0.8\textwidth]{diagrams/sweep/sweep23src.jpg}}\\
  \small{Target Context $o_0$} \hspace*{0.45cm}
  \vcenteredinclude{\includegraphics[width=0.145\textwidth]{diagrams/sweep/sweep23ctx.png}}\\
  \small{Translated Video} \hspace*{0.3cm}
  \vcenteredinclude{\includegraphics[width=0.8\textwidth]{diagrams/sweep/sweep23trans.jpg}}\\
  \caption{Example illustrations of demonstrations for a sweeping task (top) being performed in a new context (middle), with the translated observation sequences (bottom).}
\end{figure}

\newpage
\subsection*{Striking Simulation}
\begin{figure}[h!]
  \centering
  \includegraphics[width=0.8\textwidth]{diagrams/strike/strike1.jpg}
  \includegraphics[width=0.8\textwidth]{diagrams/strike/strike2.jpg}
  \caption{Example expert training demonstrations from different viewpoints.}
\end{figure}

\begin{figure}[h]
  \small{Source Video} \hspace*{0.73cm}
  \vcenteredinclude{
  \includegraphics[width=0.8\textwidth]{diagrams/strike/strikesrc1.jpg}}\\
  \small{Target Context $o_0$} \hspace*{0.45cm}
  \vcenteredinclude{\includegraphics[width=0.145\textwidth]{diagrams/strike/strikectx1.png}}\\
  \small{Translated Video} \hspace*{0.3cm}
  \vcenteredinclude{\includegraphics[width=0.8\textwidth]{diagrams/strike/striketrans1.jpg}}\\
  
  \small{Source Video} \hspace*{0.73cm}
  \vcenteredinclude{
  \includegraphics[width=0.8\textwidth]{diagrams/strike/strikesrc2.jpg}}\\
  \small{Target Context $o_0$} \hspace*{0.45cm}
  \vcenteredinclude{\includegraphics[width=0.145\textwidth]{diagrams/strike/strikectx2.png}}\\
  \small{Translated Video} \hspace*{0.3cm}
  \vcenteredinclude{\includegraphics[width=0.8\textwidth]{diagrams/strike/striketrans2.jpg}}\\
  \caption{Example illustrations of demonstrations for a striking task (top) being performed in a new context (middle), with the translated observation sequences (bottom).}
\end{figure}